\documentclass[letterpaper]{article} 
\usepackage{aaai2027}  
\usepackage[hyphens]{url}  
\usepackage{graphicx} 
\usepackage{natbib}  
\usepackage{caption} 
\usepackage{algorithm}

\usepackage{amsmath}
\usepackage{amssymb}

\usepackage{booktabs}
\usepackage{multirow}
\usepackage{graphicx}
\usepackage[table]{xcolor}

\definecolor{bestgreen}{RGB}{169, 208, 142}
\definecolor{secondgreen}{RGB}{226, 239, 218}

\newcommand{\best}[1]{\cellcolor{bestgreen}\textbf{#1}}
\newcommand{\second}[1]{\cellcolor{secondgreen}#1}

\usepackage{algpseudocode}

\usepackage{booktabs}
\usepackage{multirow}
\usepackage{graphicx}
\usepackage{array}
\usepackage[table]{xcolor}

\usepackage{tabularx}
\usepackage{array}

\newcolumntype{Y}{>{\centering\arraybackslash}X}

\setkeys{Gin}{draft=false}

\definecolor{bestgreen}{HTML}{A9D18E}
\definecolor{secondgreen}{HTML}{E2F0D9}

\newcolumntype{C}[1]{>{\centering\arraybackslash}m{#1}}

\usepackage{newfloat}
\usepackage{listings}
\DeclareCaptionStyle{ruled}{labelfont=normalfont,labelsep=colon,strut=off} 
\floatstyle{ruled}
\newfloat{listing}{tb}{lst}{}
\floatname{listing}{Listing}

\usepackage{booktabs}

\title{SkillTrace: Traversing a Query–Skill Graph for Composable LLM Agents}
\author{
    Yue Yao\textsuperscript{\rm 1},
    Shengyuan Wang\textsuperscript{\rm 2},
    Xin Chen\textsuperscript{\rm 1},
    Minke Zhang\textsuperscript{\rm 1},
    Jia He\textsuperscript{\rm 1},
    Bingjun Luo\textsuperscript{\rm 3}
\thanks{Corresponding author: bingjunluo@outlook.com},
    Tom Gedeon\textsuperscript{\rm 4}
}
\affiliations{
    \textsuperscript{\rm 1}Shandong University,\\
    \textsuperscript{\rm 2}The Australian National University,\\
    \textsuperscript{\rm 3}Tsinghua University,\\
    \textsuperscript{\rm 4}Curtin University\\
}

\begin{document}

\maketitle

\begin{abstract}

Large language model agents increasingly solve complex tasks by composing reusable skills from a library. To address this, the key challenge is not merely to retrieve individually relevant skills, but to identify a complete and executable skill composition. In this paper, we argue that this problem can be solved in a graph with three levels: compositional relations among skill queries, similarity between queries and candidates in the skill library, and the dependencies among the selected candidates. We introduce SkillTrace, which organizes the user query into a semantic hierarchy, matches skill queries and candidates, and propagates over the skill dependencies. Experiments on SkillsBench and ALFWorld demonstrate that SkillTrace achieves state-of-the-art performance, reaching a success rate of 53.17\% on SkillsBench and 91.43\% on ALFWorld. SkillTrace also delivers consistent improvements across different backbone language models, demonstrating the generality and robustness of graph-based skill retrieval.

\end{abstract}


\section{Introduction}

Large language model agents have shown increasing promise in solving complex tasks through reasoning, planning, and interaction with external environments. A growing line of work equips these agents with reusable skills, where each skill encapsulates a task-specific capability, executable procedure, or tool-use routine that can be invoked when needed~\cite{wang2023voyager}. For example, SkillsBench represents skills as structured packages of procedural knowledge, including instructions, scripts, and resources for completing expertise-intensive workflows~\cite{li2026skillsbench}. ToolBench evaluates API-based capabilities, where an agent must select appropriate tools and compose them into a valid sequence of calls to fulfill single- or multi-tool instructions~\cite{qin2023toolllm}. In embodied environments such as ALFWorld, reusable capabilities can take the form of high-level policies~\cite{shridhar2020alfworld}. Despite their different forms, these benchmarks share a common requirement: an agent is required to identify and coordinate the capabilities needed to complete a complex task.

\begin{figure}[t]
    \centering
    \includegraphics[width=\columnwidth]{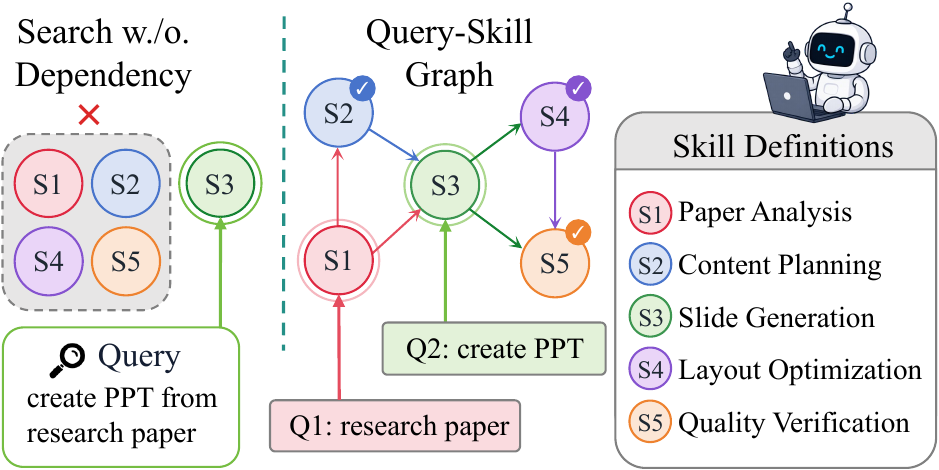}
    \caption{\textbf{Motivation.} We argue that independent skill retrieval may overlook skill dependencies, while SkillTrace connects atomic queries and skills through a query–skill graph, enabling dependency-aware traversal to retrieve a composable skill set.}
    \label{fig:motivation}
    \vspace{-1.5em}
\end{figure}

Unlike conventional information retrieval~\cite{karpukhin2020dense}, retrieval-augmented generation~\cite{lewis2020retrieval}, and tool selection~\cite{patil2024gorilla,qin2023toolllm}, composable skill retrieval requires identifying a coherent set of skills that jointly enable task execution. This setting introduces three key challenges. First, a complex user query should be decomposed into atomic skill queries while preserving their compositional relations. Second, matching skill queries to candidate skills requires both semantic alignment in task functionality and compatibility between their input–output specifications. Third, the selected skills should be completed according to their dependencies, since prerequisite or complementary skills may not be explicitly specified in the original query. These challenges motivate a unified graph formulation that jointly models relations among skill queries, connections between skill queries and candidate skills, and dependencies among skills.


In this paper, we argue that this challenge can be naturally formulated as graph search. As shown in Fig.~\ref{fig:motivation}, specifically, according to the challenge mentioned above, skill retrieval also involves three levels of relations. First, a complex query can be decomposed into finer-grained skill queries that are compositionally related to one another. These relations preserve how atomic requirements jointly constitute the original task. Second, each skill query is semantically connected to candidate skills in the library, reflecting which skill can best fulfill a particular requirement. Third, candidate skills are themselves connected through dependency relations, since executing one skill may require prerequisite or complementary skills that are not explicitly mentioned in the user query. Together, these relations form a Query–Skill Graph that connects the internal structure of the query with the organization of the skill library. Searching this graph provides a principled way to move beyond isolated semantic retrieval and identify a coherent skill composition.

\begin{figure*}[t]
    \centering
    \includegraphics[width=\textwidth]{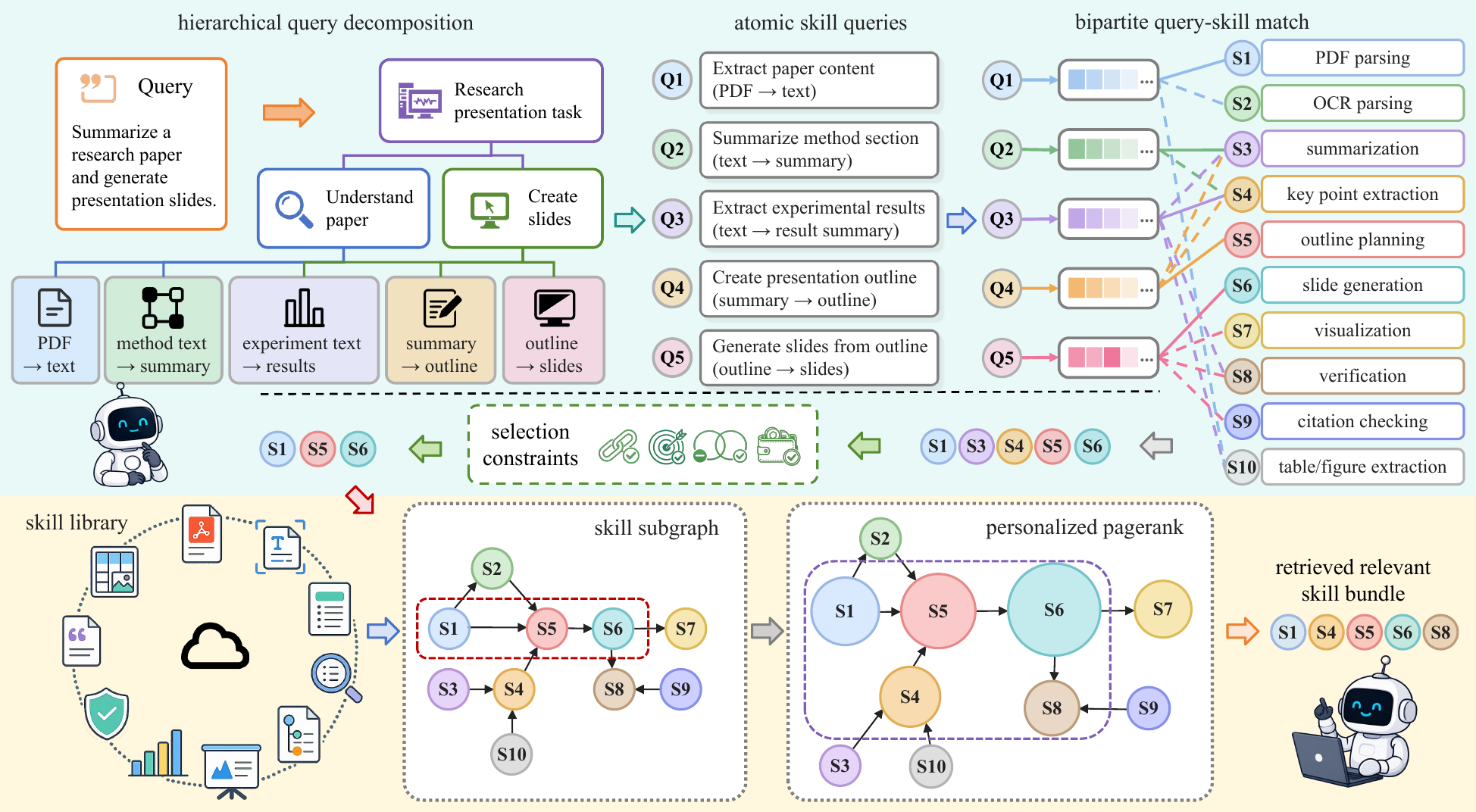}
    \caption{\textbf{Overview of SkillTrace with an intuitive example.} SkillTrace decomposes a complex query into atomic skill queries, performs bipartite query–skill matching to identify seed skills, and traverses the skill subgraph with personalized PageRank and selection constraints to retrieve a composable skill bundle.}
    \label{fig:method}
\end{figure*}

We evaluate SkillTrace on two well-established benchmarks, \emph{i.e.}, SkillsBench~\cite{li2026skillsbench} and ALFWorld~\cite{shridhar2020alfworld}, covering cross-domain procedural task solving and embodied multi-step interaction. SkillTrace achieves state-of-the-art performance on both benchmarks, reaching 53.17\% success rate on SkillsBench and 91.43\% success rate on ALFWorld. It also consistently improves different backbone language models, demonstrating the generality of the proposed Query–Skill Graph formulation. These results further validate the benefit of explicitly structuring skill retrieval as a graph, which enables the agent to jointly exploit query composition, query–skill correspondence, and inter-skill dependencies when identifying executable skill compositions.


\section{Related Work}

\textbf{Composable skills for LLM agents.} LLM agents increasingly rely on reusable skills to solve complex tasks. A skill packages procedural knowledge for a specific capability. It may include instructions, executable scripts, reference materials, and task-solving routines. Skills can be reused across tasks rather than generated from scratch for each request. More importantly, complex tasks often require several skills to work together. These skills may cover different subtasks, depend on one another, or need to be applied in a particular order. Effective skill augmentation therefore requires more than retrieving a single relevant skill. The agent needs to identify a suitable skill set and use its members coherently during execution. SkillsBench shows that skill utility varies across tasks and that poorly matched skills can reduce agent performance~\cite{li2026skillsbench}. SRA-Bench further divides skill augmentation into retrieval, incorporation, and execution, showing that successful retrieval does not guarantee successful use~\cite{su2026skill}. Skills in the Wild studies large and noisy skill repositories, where distractors and overly general skills make suitable capabilities harder to identify and combine~\cite{liu2026well}.

\textbf{Skill retrieval methods.} As skill libraries grow, placing all available skills in the prompt becomes increasingly ineffective. Recent methods therefore retrieve a compact set of task-relevant skills before execution. SkillRet improves query--skill matching through retrieval-specific training, while SkillRouter adopts a retrieve-and-rerank pipeline that considers more complete skill content~\cite{cho2026skillret,zheng2026skillrouter}. However, these methods largely score skills as independent candidates and return a ranked top-$k$ list. Such a list can identify individually relevant skills, but it cannot determine whether they depend on one another, conflict with each other, provide redundant functionality, or require a specific execution order. To move beyond independent skill ranking, graph-based methods explicitly model relations among skills. SkillNet organizes skills with an ontology and a relation graph, while AgentSkillOS retrieves skills through a capability tree and composes them into task-specific directed acyclic graphs~\cite{liang2026skillnet,li2026organizing}. Graph-of-Skills further uses graph propagation to expand initially relevant skills with their prerequisites~\cite{liu2026graph}. SkillGraph and SkillDAG incorporate execution experience to update skill nodes and relations~\cite{li2026skillgraph,bai2026skilldag}. Nevertheless, constructing a skill graph does not directly determine which connected skills should be selected together for a specific task. Existing methods often rely on fixed propagation, predefined update rules, or LLM-based orchestration, rather than jointly learning a task-specific executable subgraph under dependency and context-budget constraints.

\section{Method}

\subsection{Problem Definition}

Our goal is to retrieve a set of skills from the skill library for a given complex query, ensuring that the selected skills cover the query requirements and include the dependencies needed for successful execution. This approach enables an LLM agent to complete the target task with a complete and executable skill composition. In order to retrieve the desired skill set effectively, we propose the SkillTrace framework. 

Given a user query $x$ and a skill library $\mathcal{S}=\{s_1,\ldots,s_M\}$, our goal is to retrieve a subset of skills that can be composed to accomplish the task specified by $x$.
Let selected skill subset $\mathcal{S}'\subseteq\mathcal{S}$, and let $\mathcal{R}(x,\mathcal{S}')$ measure the resulting task success rate. The objective of composable skill retrieval is formulated as

\begin{equation}
\mathcal{S}_{x}^{*}
=
\operatorname*{arg\,max}_{\mathcal{S}'\subseteq\mathcal{S}}
\mathcal{R}
\left(
x,\mathcal{S}'
\right).
\label{eq:skill_retrieval_objective}
\end{equation}

Unlike independent skill retrieval, the utility of a skill subset is determined jointly by its constituent skills. The selected skills should collectively cover the requirements of the query, match the corresponding task content and input-output specifications, and include the dependencies required for execution. Importantly, retrieving more skills does not necessarily improve task performance, since irrelevant, redundant, or incompatible skills may interfere with skill composition and agent execution.

Directly optimizing Eq.~\eqref{eq:skill_retrieval_objective} is usually not practical as evaluating a candidate skill composition requires executing the complete downstream task, which is often time-consuming and computationally expensive. Moreover, task performance can be observed only after execution and generally provides no tractable differentiable objective for directly optimizing the discrete skill subset. Exhaustively evaluating possible skill combinations is also infeasible as the search space (\emph{i.e.}, skill library size) grows exponentially with the size of the skill library. We therefore formulate composable skill retrieval as a structured search problem and develop SkillTrace to efficiently identify promising skill compositions without repeatedly executing the downstream task.

\subsection{Query-Skill Graph Construction.}

As shown in Fig.~\ref{fig:method}, SkillTrace instead addresses this objective through a structured graph traversal process. Given an input query $x$ and a skill library $\mathcal{S}$, SkillTrace first decomposes $x$ into atomic skill queries, then identifies a primary skill for 
each atomic requirement through query--skill assignment, and finally traverses 
the dependencies among skills to retrieve the supporting skills required for 
execution. We organize these relations into a Query-Skill Graph:

\begin{equation}
\mathcal{G}_{x,\mathcal{S}} = \mathcal{T}_{x} \cup \mathcal{B}_{x,\mathcal{S}} \cup \mathcal{D}_{\mathcal{S}},
\label{eq:query_skill_graph}
\end{equation}
with $\mathcal{T}_{x}$, $\mathcal{B}_{x,\mathcal{S}}$, and 
$\mathcal{D}_{\mathcal{S}}$ being the hierarchical query tree, the 
query-skill bipartite graph, and the skill dependency graph, respectively. 
Here, $\mathcal{T}_{x}$ is determined by the input query $x$, 
$\mathcal{B}_{x,\mathcal{S}}$ jointly depends on the query $x$ and the skill 
library $\mathcal{S}$, and $\mathcal{D}_{\mathcal{S}}$ represents the global 
dependency structure defined over $\mathcal{S}$.

\paragraph{Hierarchical query tree.}
Given an input query $x$, we prompt a language model $p_{\theta}$ to organize
its requirements into a hierarchical semantic tree. Formally, the tree is
generated by
\begin{equation}
\mathcal{T}_{x}
\sim
p_{\theta}
\left(
\mathcal{T}
\mid
x,
p_{\mathrm{atom}}
\right),
\label{eq:query_tree_generation}
\end{equation}
where $p_{\mathrm{atom}}$ is the atomic decomposition prompt that specifies
the output schema and the atomicity constraints. Internal nodes represent
higher-level actions that group related requirements, whereas each leaf
represents a self-contained atomic skill query. The atomic skill query set is
obtained as $
\mathcal{A}_{x}
=
\operatorname{Leaf}
\left(
\mathcal{T}_{x}
\right).$
Each atomic skill query specifies one task requirement, its input and output
types, and a rationale for why it can be primarily fulfilled by one skill.
The prompt requires each leaf to describe exactly one input--output
transformation, prohibits decomposition below the skill level, and prevents
the introduction of requirements absent from the original query.

\paragraph{Query-Skill bipartite graph.}
Given the atomic skill queries $\mathcal{A}_{x}$ and the skill library
$\mathcal{S}$, SkillTrace constructs a weighted query--skill bipartite graph

\begin{equation}
\mathcal{B}_{x,\mathcal{S}}
=
\left(
\mathcal{A}_{x},
\mathcal{S},
\mathcal{E}^{B}_{x,\mathcal{S}},
\mathcal{W}^{B}_{x,\mathcal{S}}
\right),
\label{eq:query_skill_bipartite_graph}
\end{equation}
where $\mathcal{A}_{x}$ and $\mathcal{S}$ are two disjoint node sets,
$\mathcal{E}^{B}_{x,\mathcal{S}}
\subseteq \mathcal{A}_{x}\times\mathcal{S}$ denotes the query--skill edges,
and $\mathcal{W}^{B}_{x,\mathcal{S}}$ denotes their semantic compatibility
scores. Specifically, the edge weight between an atomic skill query
$q_i\in\mathcal{A}_{x}$ and a skill $s_j\in\mathcal{S}$ is defined as
$
w_{ij}
=
\operatorname{sim}
\left(
f_Q(q_i),
f_S(s_j)
\right),
$
where $f_Q(\cdot)$ and $f_S(\cdot)$ encode atomic skill queries and skills
into a shared representation space, respectively, and
$\operatorname{sim}(\cdot,\cdot)$ denotes cosine similarity.

Since each atomic skill query is designed to be fulfilled by one primary
skill, SkillTrace selects the highest-scoring skill for each query:

\begin{equation}
s_i^{*}
=
\underset{s_j\in\mathcal{S}}{\operatorname{argmax}}
\; w_{ij},
\qquad
q_i\in\mathcal{A}_{x}.
\label{eq:primary_skill_selection}
\end{equation}
The resulting primary skill set is
$\mathcal{P}_{x,\mathcal{S}}
=
\{s_i^{*}\mid q_i\in\mathcal{A}_{x}\}$.
These primary skills directly address the atomic requirements represented
by the leaf nodes of $\mathcal{T}_{x}$ and serve as seed nodes for subsequent
dependency-aware traversal over the skill graph.

SkillTrace then performs maximum-weight bipartite matching over
$\mathcal{B}_{x,\mathcal{S}}$ to identify the primary skills. The matching
maximizes the total semantic compatibility of the selected query--skill
edges, subject to the constraint that each atomic skill query is matched
with exactly one skill and each skill is matched with at most one atomic
skill query. This one-to-one constraint prevents multiple atomic queries
from selecting the same skill and encourages the matched skills to jointly
cover the distinct atomic requirements of the input query.

The skills incident to the matched edges constitute the primary skill set
$\mathcal{P}_{x,\mathcal{S}}$. These primary skills directly address the
atomic requirements represented by the leaf nodes of $\mathcal{T}_{x}$ and
serve as seed nodes for subsequent dependency-aware traversal over the
skill graph.

\paragraph{Skill dependency subgraph.}
Primary-skill assignment alone may not produce an executable skill composition 
because a selected skill can require additional prerequisite or supporting 
skills. We therefore organize the skill library as a directed skill dependency 
graph:
\begin{equation}
\mathcal{D}_{\mathcal{S}} =
\left(
\mathcal{S},
\mathcal{E}^{D}_{\mathcal{S}},
\mathcal{W}^{D}_{\mathcal{S}}
\right),
\label{eq:skill_dependency_graph}
\end{equation}
where $\mathcal{S}$ denotes the skill-node set, 
$\mathcal{E}^{D}_{\mathcal{S}}\subseteq\mathcal{S}\times\mathcal{S}$ denotes 
the directed dependency-edge set, and $\mathcal{W}^{D}_{\mathcal{S}}$ contains 
the corresponding dependency weights. A directed edge 
$(s_i,s_j)\in\mathcal{E}^{D}_{\mathcal{S}}$ indicates that executing skill 
$s_i$ requires the capability, state, or output provided by skill $s_j$.

A directed edge
$(s_i,s_j)\in\mathcal{E}^{D}_{\mathcal{S}}$
indicates that the output of skill $s_i$ can be consumed by skill $s_j$.
The dependency edges and their weights are defined as
\begin{equation}
\begin{aligned}
w^{D}_{ij}
&=
\operatorname{overlap}
\left(
\mathcal{O}_i,
\mathcal{I}_j
\right),\\
\mathcal{E}^{D}_{\mathcal{S}}
&=
\left\{
(s_i,s_j)
\mid
w^{D}_{ij}
\geq
\delta_D
\right\},
\end{aligned}
\label{eq:skill_dependency_construction}
\end{equation}
where $\mathcal{O}_i$ and $\mathcal{I}_j$ denote the output and input schema
of skills $s_i$ and $s_j$, respectively. Overlap function measures whether input and output are matched. 

\paragraph{Skill augmentation.}
SkillTrace follows the Reverse Personalized PageRank (ReversePPR) paradigm~\cite{yang2024efficient} to augment the
primary skill set with structurally relevant supporting skills. Specifically,
we treat $\mathcal{P}_{x,\mathcal{S}}$ as personalized seed nodes and propagate
their relevance through the skill dependency graph. The query--skill matching
scores of the primary skills are normalized to form the personalization vector
$\mathbf{p}$, whose nonzero entries correspond to
$\mathcal{P}_{x,\mathcal{S}}$.

Let $\mathbf{A}^{D}$ denote the weighted adjacency matrix of
$\mathcal{D}_{\mathcal{S}}$, where
$A^{D}_{ij}=w^{D}_{ij}$ if
$(s_i,s_j)\in\mathcal{E}^{D}_{\mathcal{S}}$, and
$A^{D}_{ij}=0$ otherwise. Since a dependency edge is directed from a
supporting skill to the skill that consumes its output, we additionally
include reverse transitions to propagate relevance from a primary skill
toward its prerequisites. The resulting reverse-aware PPR process is

\begin{equation}
\begin{aligned}
&\mathbf{T}^{D}
=
\operatorname{RowNorm}
\left(
\mathbf{A}^{D}
+
(\mathbf{A}^{D})^{\top}
\right),\\
&\mathbf{r}^{(t+1)}
=
\alpha\mathbf{p}
+
(1-\alpha)
(\mathbf{T}^{D})^{\top}
\mathbf{r}^{(t)}.
\end{aligned}
\label{eq:dependency_aware_propagation}
\end{equation}

Here, $\operatorname{RowNorm}(\cdot)$ denotes row normalization,
$\mathbf{T}^{D}$ is the reverse-aware dependency transition matrix, and
$\mathbf{r}^{(t)}$ is the skill relevance vector at iteration $t$, initialized
as $\mathbf{r}^{(0)}=\mathbf{p}$. The restart probability
$\alpha\in(0,1)$ controls the balance between retaining relevance on the
primary skills and propagating it through their dependency neighborhoods.
The iteration continues until convergence, yielding the stationary relevance
vector $\mathbf{r}^{*}$.

Finally, SkillTrace retains the primary skills and augments them with the
$K$ highest-scoring supporting skills:
\begin{equation}
\mathcal{S}_{x}^{*}
=
\mathcal{P}_{x,\mathcal{S}}
\cup
\operatorname{TopK}_{\,
s_i\in
\mathcal{S}
\setminus
\mathcal{P}_{x,\mathcal{S}}
}
\left(
r_i^{*}
\right),
\label{eq:final_skill_set}
\end{equation}
where $K$ specifies the maximum number of supporting skills added through
dependency-aware propagation.

Algorithm~\ref{alg:skilltrace} summarizes the complete traversal of the 
Query-Skill Graph. Overall, SkillTrace first traverses query composition through 
$\mathcal{T}_{x}$, then resolves query--skill correspondence through 
global bipartite matching on $\mathcal{B}_{x}$, and finally recovers 
execution dependencies through reverse-aware propagation on 
$\mathcal{G}_{S}$. This staged traversal converts a complex query into a 
query-aligned and dependency-aware skill composition.


\begin{algorithm}[t]
  \caption{SkillTrace for Composable Skill Retrieval}
  \label{alg:skilltrace}
  \begin{algorithmic}[1]
    \State \textbf{Input:} query $x$, skill library $\mathcal{S}$,
    dependency threshold $\delta_D$, number of supporting skills $K$,
    and restart probability $\alpha$.
    \State \textbf{Begin:}

    \State $\mathcal{D}_{\mathcal{S}}
    \gets
    (\mathcal{S},\emptyset,\emptyset)$
    \Comment{dependency graph init.}

    \For{each ordered pair $(s_i,s_j)\in\mathcal{S}\times\mathcal{S}$}
        \State $w^{D}_{ij}\gets
        \operatorname{overlap}(\mathcal{O}_i,\mathcal{I}_j)$
        \If{$w^{D}_{ij}\geq\delta_D$}
            \State Add $(s_i,s_j,w^{D}_{ij})$
            to $\mathcal{D}_{\mathcal{S}}$
        \EndIf
    \EndFor
    \Comment{dependency construction}

    \State $\mathcal{T}_{x}\sim
    p_{\theta}(\mathcal{T}\mid x,p_{\mathrm{atom}})$,
    $\mathcal{A}_{x}\gets\operatorname{Leaf}(\mathcal{T}_{x})$
    \Comment{query decomposition}

    \State $\mathcal{B}_{x,\mathcal{S}}
    \gets
    \{w_{ij}=
    \operatorname{sim}(f_Q(q_i),f_S(s_j))\}$
    \Comment{query--skill graph}

    \State $\sigma^{*}\gets
    \operatorname{Hungarian}(\mathcal{B}_{x,\mathcal{S}})$
    \Comment{primary matching}

    \State $\mathcal{P}_{x,\mathcal{S}}
    \gets
    \{s_{\sigma^{*}(i)}
    \mid q_i\in\mathcal{A}_{x}\}$

    \State $\mathbf{p}\gets
    \operatorname{Normalize}
    (\{w_{i,\sigma^{*}(i)}\})$
    \Comment{seed initialization}

    \State $\mathbf{r}^{*}\gets
    \operatorname{ReversePPR}
    (\mathcal{D}_{\mathcal{S}},\mathbf{p},\alpha)$
    \Comment{dependency propagation}

    \State $\mathcal{S}_{x}^{*}
    \gets
    \mathcal{P}_{x,\mathcal{S}}
    \cup
    \operatorname{TopK}_{\,
    \mathcal{S}\setminus\mathcal{P}_{x,\mathcal{S}}
    }
    (\mathbf{r}^{*})$
    \Comment{skill augmentation}

    \State \Return $\mathcal{S}_{x}^{*}$
  \end{algorithmic}
\end{algorithm}

\section{Experiment}

We evaluate whether traversing the Query-Skill Graph improves task performance, skill retrieval quality, and execution efficiency compared with full-library access and existing skill retrieval methods.

\begin{table*}[t]
  \caption{\textbf{Main results on SkillsBench and ALFWorld.} We show the superiority of SkillTrace over existing methods. Core, Extended, and Extreme are the three subsets divided in SkillsBench, while Easy, Medium, and Hard group ALFWorld tasks according to the number of required atomic skills. Category columns report the number of successful tasks, and SR reports the overall success rate (\%) on SkillsBench or ALFWorld. The best and second-best results in each column are highlighted in dark green and light green, respectively.}
  \vspace{-0.5em}
  \label{tab:main_results}

  \centering
  \footnotesize
  \setlength{\tabcolsep}{4.5pt}
  \renewcommand{\arraystretch}{1.12}

  \begin{tabularx}{0.90\textwidth}{
    l
    *{8}{Y}
  }
    \toprule

    \multicolumn{1}{c}{%
      \multirow[c]{2}{*}{\textbf{Method}}
    }
    &
    \multicolumn{4}{c}{\textbf{SkillsBench}}
    &
    \multicolumn{4}{c}{\textbf{ALFWorld}}
    \\

    \cmidrule(lr){2-5}
    \cmidrule(lr){6-9}

    &
    Core
    &
    Extended
    &
    Extreme
    &
    SR (\%)
    &
    Easy
    &
    Medium
    &
    Hard
    &
    SR (\%)
    \\

    \midrule

    Vanilla Skills
    &
    2
    &
    17
    &
    10
    &
    36.69
    &
    \second{38}
    &
    \best{68}
    &
    \second{20}
    &
    \second{90.00}
    \\

    Vector Skills
    &
    \second{4}
    &
    22
    &
    10
    &
    42.12
    &
    37
    &
    65
    &
    \best{21}
    &
    87.86
    \\

    SkillDAG~\cite{bai2026skilldag}
    &
    3
    &
    20
    &
    \second{12}
    &
    41.38
    &
    37
    &
    \best{68}
    &
    \best{21}
    &
    \second{90.00}
    \\

    GoS~\cite{liu2026graph}
    &
    3
    &
    \second{24}
    &
    \second{12}
    &
    \second{46.49}
    &
    36
    &
    \second{66}
    &
    19
    &
    86.43
    \\

    SkillTrace
    &
    \best{5}
    &
    \best{26}
    &
    \best{13}
    &
    \best{53.17}
    &
    \best{39}
    &
    \best{68}
    &
    \best{21}
    &
    \best{91.43}
    \\

    \bottomrule
  \end{tabularx}
  \vspace{-0.7em}
\end{table*}

\subsection{Experiment Setup}

\textbf{Benchmarks.} We evaluate SkillTrace on two benchmarks, SkillsBench~\cite{li2026skillsbench} and ALFWorld~\cite{shridhar2020alfworld}. 
SkillsBench contains expertise-intensive technical tasks spanning eight domains, with skills represented as structured packages of instructions, scripts, templates, and reference materials~\cite{li2026skillsbench}. Many tasks require multiple skills to be identified and combined into a complete workflow. We conduct experiments on its Core, Extended, and Extreme subsets. 
ALFWorld is a text-based embodied benchmark, and we evaluate 140 tasks from its development in-distribution split. These tasks involve multi-step household interactions, including navigation, object search, and object manipulation~\cite{shridhar2020alfworld}. For analysis, we stratify the 140 ALFWorld episodes by the number of atomic skill queries identified through hierarchical query decomposition. We use this number as a proxy for compositional difficulty: more atomic skill queries indicate that a task contains more distinct requirements and therefore requires the agent to retrieve, coordinate, and execute a larger set of skills. Accordingly, episodes with one atomic skill query are categorized as Easy (43 episodes), those with two as Medium (73 episodes), and those with three or more as Hard (24 episodes). Together, the two benchmarks evaluate SkillTrace on both cross-domain technical problem solving and sequential embodied interaction.

\textbf{Baselines.}
We compare SkillTrace with four baselines.
\textit{Vanilla Skills} provides the entire skill library to the agent without retrieval, allowing full access to all skills but introducing a large and potentially noisy context.
\textit{Vector Skills} retrieves a fixed number of skills according to the semantic similarity between the original query and each skill description. It treats skills independently and does not consider their dependencies.
\textit{SkillDAG} organizes skills and their relations as a directed acyclic graph, enabling graph-based skill selection while retaining the original query as a single retrieval unit~\cite{bai2026skilldag}.
\textit{GoS} first retrieves query-relevant seed skills and then propagates their relevance over a skill graph to identify additional skills connected through structural relations~\cite{liu2026graph}.

\textbf{Models and evaluation metrics.} For the main comparison, we use GPT-5.6 as the backbone language model for all methods, so that the performance differences primarily reflect the effectiveness of different skill retrieval strategies. On SkillsBench, we report the number of successfully completed tasks on the Core, Extended, and Extreme subsets, together with the overall success rate $SR$ (\%). On ALFWorld, we report the number of successfully completed tasks in the Easy, Medium, and Hard groups, together with the overall success rate $SR$ (\%). Higher values indicate better task performance.

Since the effectiveness of a skill retrieval method may vary across backbone models, we further evaluate SkillTrace with DeepSeek-V3.2~\cite{liu2025deepseek}, Kimi-K2.5~\cite{team2026kimi}, Qwen3.5-397B-A17B~\cite{qwen2026qwen35}, and Gemini 3.1 Flash-Lite~\cite{google2026gemini31flashlite}. In this cross model evaluation, we compare SkillTrace with GoS under the same skill library, task instructions, retrieval budget, and execution environment.

\subsection{Main Results}

\textbf{SkillTrace achieves the strongest overall performance on both benchmarks.} As shown in Table~\ref{tab:main_results}, SkillTrace achieves the highest overall success rate of 53.17\% on SkillsBench and the highest success rate of 91.43\% on ALFWorld. On SkillsBench, it outperforms GoS, the strongest baseline, by 6.68 percentage points, and exceeds SkillDAG, Vector Skills, and Vanilla Skills by 11.79, 11.05, and 16.48 points, respectively. These substantial gains over both graph-based and similarity-based baselines indicate that semantic relevance or dependency modeling alone is insufficient; effective skill retrieval requires jointly covering the query requirements and completing the dependencies needed for execution.

On ALFWorld, SkillTrace surpasses SkillDAG, the strongest baseline, by 1.43 percentage points and improves over GoS and Vector Skills by 5.00 and 3.57 points, respectively. The consistent improvements across SkillsBench and ALFWorld demonstrate that SkillTrace generalizes well to substantially different settings, ranging from expertise-intensive technical tasks to long-horizon embodied interactions. This suggests that its query decomposition, query--skill matching, and dependency-aware traversal together produce skill compositions that are not only relevant but also complete and executable.

\textbf{The overall improvements are consistent across task subsets.} The advantage of SkillTrace is not driven by a single task category. On SkillsBench, it achieves the best results on Core, Extended, and Extreme, successfully completing 5, 26, and 13 tasks, respectively. Compared with the strongest baseline in each subset, SkillTrace solves one more Core task, two more Extended tasks, and one more Extreme task. On ALFWorld, SkillTrace successfully completes 39 Easy, 68 Medium, and 21 Hard tasks. It achieves the best result on Easy and matches the strongest results on Medium and Hard. The consistent category-wise performance shows that SkillTrace remains effective across tasks requiring different numbers and combinations of skills.

\textbf{Task-specific graph traversal is more effective than flat retrieval and graph organization alone.} On SkillsBench, Vanilla Skills provides the entire skill library to the agent and obtains an overall success rate of 36.69\%, while Vector Skills independently retrieves skills according to semantic similarity and reaches 42.12\%. SkillDAG obtains 41.38\%, indicating that organizing skills into a graph does not by itself guarantee improved task performance. GoS increases the success rate to 46.49\% by propagating relevance over skill relations, while SkillTrace further improves it to 53.17\% by aligning atomic task requirements with primary skills and recovering their dependencies through graph traversal. A similar advantage is observed on ALFWorld, where SkillTrace achieves 91.43\%, outperforming all flat and graph-based baselines. These results show that effective skill retrieval requires not only modeling skill relations, but also traversing them according to the specific requirements of each query. 

\begin{figure}[t]
    \centering
    \includegraphics[width=\columnwidth]{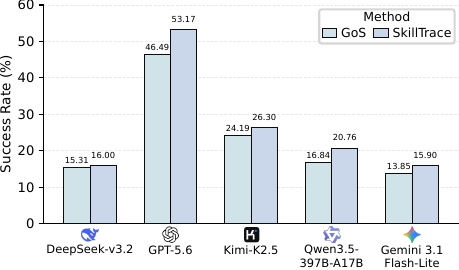}
    \vspace{-1em}
    \caption{Success rate comparison between GoS and SkillTrace across five backbone language models on SkillsBench.}
    \label{fig:cross_model_reward}
    \vspace{-1em}
\end{figure}

\subsection{Cross Model Generalization}

In the Table~\ref{tab:main_results}, we use GPT-5.6 as the common backbone model to control the influence of model capability and focus the comparison on different skill retrieval methods. However, results obtained with a single backbone cannot determine whether the improvement of SkillTrace comes from its retrieval mechanism or from the particular reasoning and execution capabilities of GPT-5.6. We therefore conduct an additional cross model experiment to examine whether SkillTrace remains effective when paired with backbone models of different capability levels. Specifically, we compare SkillTrace with GoS using five language models under the same skill library, retrieval budget, task instructions, and execution environment.

As shown in Fig.~\ref{fig:cross_model_reward}, SkillTrace consistently outperforms GoS across five backbone models. It improves the success rate by 0.69 points on DeepSeek-V3.2, 6.68 points on GPT-5.6, 2.11 points on Kimi-K2.5, 3.92 points on Qwen3.5-397B-A17B, and 2.05 points on Gemini 3.1 Flash-Lite. Although the magnitude of improvement varies across models, the gain remains positive in every case. This result indicates that the advantage of SkillTrace is not specific to GPT-5.6 and can generalize to backbones with different reasoning, composition, and execution capabilities.

The overall success rate nevertheless varies substantially across backbone models. With SkillTrace, GPT-5.6 achieves a success rate of 53.17\%, whereas DeepSeek-V3.2, Kimi-K2.5, Qwen3.5-397B-A17B, and Gemini 3.1 Flash-Lite obtain 16.00\%, 26.30\%, 20.76\%, and 15.90\%, respectively. This variation suggests that skill retrieval determines whether relevant and composable skills are provided, while the backbone model still affects how accurately these skills are understood and executed.

\begin{figure}[t]
    \centering
    \includegraphics[width=\columnwidth]{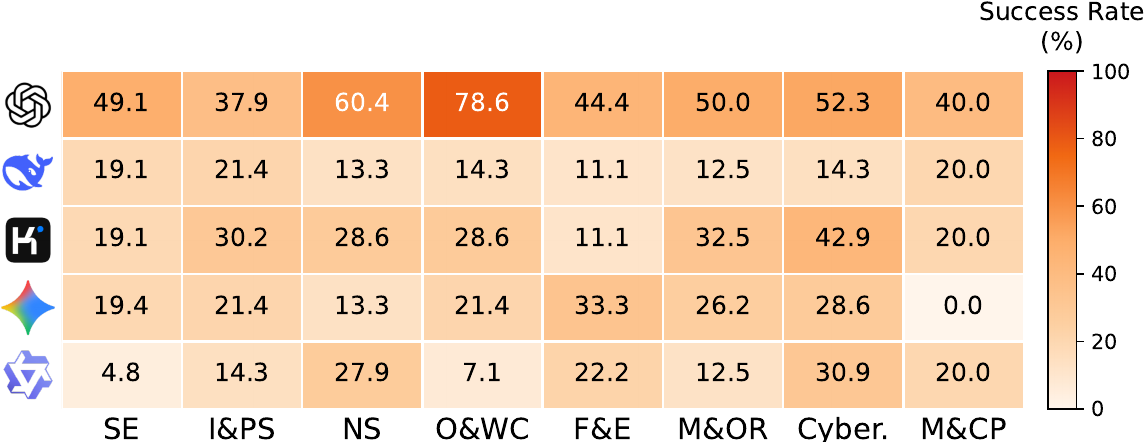}
    \caption{Category-wise performance of SkillTrace with five backbone language models across the eight SkillsBench task domains. The full names of the category abbreviations are provided in the appendix.}
    \label{fig:cross_model_category}
    \vspace{-1em}
\end{figure}

To further examine whether these differences are uniformly distributed across task domains, we report the category-wise performance of SkillTrace in Fig.~\ref{fig:cross_model_category}. GPT-5.6 achieves the highest performance across all eight SkillsBench categories, with particularly strong results in Office \& White Collar and Natural Science. The remaining models exhibit more pronounced domain-specific variation. Kimi-K2.5 performs relatively well in Industrial \& Physical Systems, Mathematics \& OR, and Cybersecurity, while Gemini 3.1 Flash-Lite obtains its strongest relative result in Finance \& Economics. These results show that different backbones retain distinct domain strengths and weaknesses, while the consistent improvements over GoS demonstrate that SkillTrace provides a generally effective retrieval mechanism across models.

\begin{figure*}[t]
    \centering
    \includegraphics[width=\textwidth]{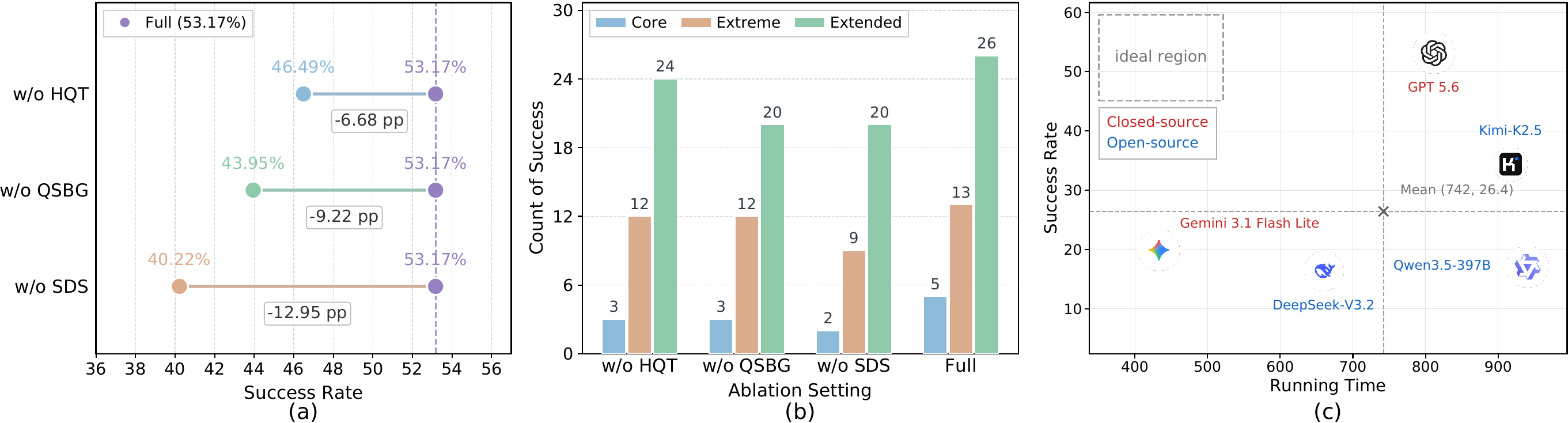}
    \vspace{-1.5em}
    \caption{\textbf{Additional analysis.} (a) Overall success rates of the full SkillTrace and its variant ablations without the Hierarchical Query Tree (\textit{w/o HQT}), Query-Skill Bipartite Graph (\textit{w/o QSBG}), or Skill Dependency Subgraph (\textit{w/o SDS}). (b) Numbers of successfully completed tasks on the Core, Extreme, and Extended subsets under each ablation setting. (c) Execution time and success rate of SkillTrace across five backbone models, where we have the success rate (\%) and the end-to-end execution time in seconds.}
    \label{fig:merge}
    \vspace{-1em}
\end{figure*}

\subsection{Ablation Study}

We conduct ablation experiments on SkillsBench to examine the contributions of the Hierarchical Query Tree (HQT), Query-Skill Bipartite Graph (QSBG), and Skill Dependency Subgraph (SDS). As shown in Fig.~\ref{fig:merge} (a) and (b), the full SkillTrace achieves an overall success rate of 53.17\%, completing 5, 26, and 13 tasks in the Core, Extended, and Extreme subsets, respectively. Without HQT, the original query is used as a single retrieval unit for primary-skill retrieval before dependency-aware augmentation. This variant achieves 46.49\%, corresponding to a decrease of 6.68 percentage points, showing that atomic query decomposition improves the coverage of distinct task requirements.

Retaining the HQT but removing the QSBG further reduces the success rate to 43.95\%, a decrease of 9.22 percentage points. Successful Extended tasks drop from 26 to 20, suggesting that structured query--skill assignment prevents redundant skill selections from becoming propagation seeds. Finally, removing the SDS and directly returning the primary skills identified by the HQT and QSBG produces the largest performance degradation. The success rate decreases to 40.22\%, while the numbers of successful Core, Extended, and Extreme tasks fall to 2, 20, and 9, respectively. This 12.95-point decrease highlights the importance of dependency-aware propagation for recovering prerequisite and supporting skills not explicitly expressed in the input query. Overall, the three components are complementary, \emph{i.e.}, the HQT represents atomic requirements, the QSBG identifies primary skills, and the SDS completes the executable skill composition.



\subsection{Efficiency Analysis}

SkillTrace models richer graph structures while maintaining manageable computational complexity. Let $N$ and $E$ denote the numbers of skills and dependency edges, respectively, $Q$ the number of atomic skill queries, and $K$ the number of propagation iterations. The skill embeddings and dependency graph are constructed offline and reused across queries. Although dependency-graph construction may require pairwise comparisons among skills, this cost is incurred only once and is amortized over subsequent tasks. At inference time, constructing the query-skill similarity matrix requires $O(QN)$ computations. Maximum-weight bipartite matching requires $O(Q^2N)$ time when $Q \leq N$, while personalized PageRank requires $O(K(N+E))$ time. Therefore, the online graph-retrieval complexity is
\begin{equation}
O\bigl(QN + Q^2N + K(N+E)\bigr).
\label{eq:online_complexity}
\end{equation}
In practice, $Q$ and $K$ are small, and the skill dependency graph is sparse. SkillTrace therefore avoids exhaustive search over skill combinations and remains approximately linear in the size of the skill library.

The end-to-end runtime results are consistent with the practical efficiency of SkillTrace. On SkillsBench, SkillTrace takes 765.00 seconds, compared with 668.80 seconds for GoS. This increase of 14.38\% is accompanied by a 6.68-point improvement. On ALFWorld, SkillTrace takes 73.66 seconds, which is 6.57\% less than GoS, while improving the success rate by 5.00 percentage points.

To examine whether stronger task performance is simply associated with longer execution time, Fig.~\ref{fig:merge} (c) compares the runtime and success rate of SkillTrace across five backbone language models. The Pearson correlation coefficient is only $r=0.08$, indicating no clear linear relationship between runtime and task success rate on this yet. For example, GPT-5.6 achieves the highest success rate without having the longest runtime, whereas Qwen3.5-397B-A17B has the longest runtime but one of the lowest success rate. Given the limited number of evaluated backbones, this result should be interpreted as descriptive evidence rather than a statistically conclusive correlation test. Overall, these results show that the performance gains of SkillTrace cannot be explained solely by increased execution time.

\section{Conclusion}

In this paper, we introduced SkillTrace, a graph-based framework for retrieving composable skills for large language model agents. SkillTrace formulates skill retrieval through three levels of relations: compositional relations among skill queries, semantic correspondences between queries and candidate skills, and dependency relations among skills. It first organizes a complex user query into a semantic hierarchy, then identifies seed skills through query--skill matching, and finally traverses the skill graph to complete the required skill composition. Experiments on SkillsBench and ALFWorld show that SkillTrace achieves state-of-the-art performance, reaching 53.17\% success rate on SkillsBench and 91.43\% success rate on ALFWorld. Its consistent improvements across different backbone language models further demonstrate the generality and robustness of the proposed formulation. These results suggest that explicitly modeling query composition and skill dependencies is important for retrieving complete and executable skill sets, providing a promising direction for building more capable skill-based language model agents.

\newpage
\bibliography{aaai2027}


\newpage
\appendix
\section{Technical Appendix}

\subsection{Dataset Details and Subset Statistics}

We use two public benchmarks with complementary forms of compositionality.
SkillsBench~\cite{li2026skillsbench} evaluates whether an agent can identify
and apply reusable skill packages to expertise-intensive technical tasks,
whereas ALFWorld~\cite{shridhar2020alfworld} evaluates sequential planning and
execution in text-based household environments. This combination lets us
measure skill retrieval in both artifact-producing technical workflows and
long-horizon embodied interaction.

\begin{figure}
    \centering
    \includegraphics[width=1\linewidth]{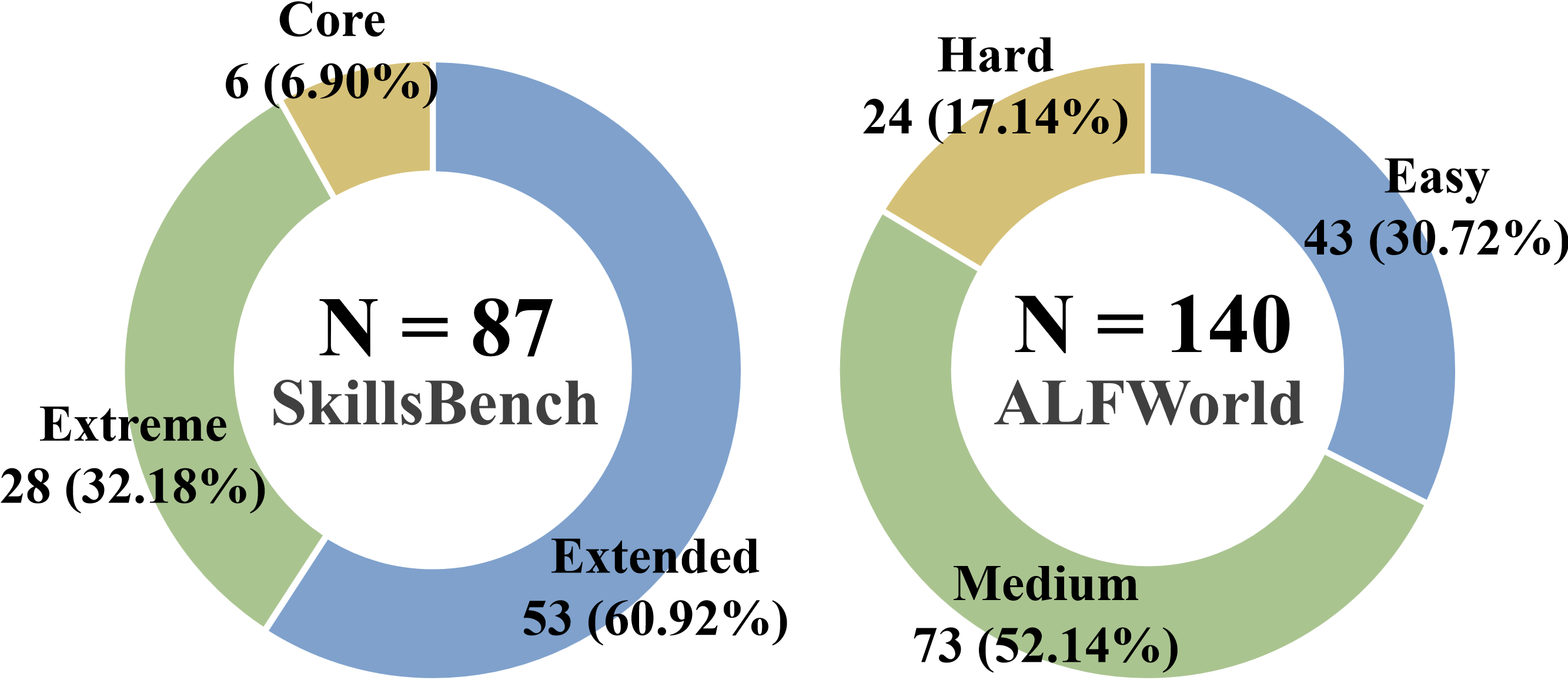}
    \caption{Benchmark subset statistics used in the main experiments.}
    \label{fig:dataset_subsets}
\end{figure}

\textbf{SkillsBench.} We evaluate all 87 tasks in the public benchmark split.
The tasks span eight domains and are grouped into the Core, Extended, and
Extreme subsets used in the main paper. The split contains 6 Core, 53
Extended, and 28 Extreme tasks. Figure~\ref{fig:dataset_subsets} reports their
exact proportions, and Table~\ref{tab:dataset_categories} reports the domain
composition.

\textbf{ALFWorld.} We evaluate all 140 episodes in the public development
in-distribution (valid\_seen) split. The official task types describe
the household objective, rather than retrieval difficulty, and comprise
Pick~\&~Place, Examine in Light, Clean~\&~Place, Heat~\&~Place,
Cool~\&~Place, and Pick Two~\&~Place. For the compositional analysis in the
main paper, we create a fixed task-to-stratum mapping from the number of
atomic skill queries in the hierarchical decomposition and reuse this mapping
for every method. Episodes containing one atomic query are labeled Easy,
those containing two are labeled Medium, and those containing three or more
are labeled Hard. These labels are an analysis stratification introduced in
this work, not official ALFWorld difficulty annotations. They measure the
number of distinct requirements that must be coordinated, rather than merely
the length of an observed action trajectory.

\begin{table}[t]
  \centering
  \small
  \caption{Task-category composition of the evaluated benchmark splits.}
  \label{tab:dataset_categories}
  \begin{tabular}{llrr}
    \toprule
    Benchmark & Category & $n$ & Share (\%) \\
    \midrule
    \multirow{8}{*}{SkillsBench}
      & Software Engineering           & 16 & 18.39 \\
      & Industrial \& Physical Systems & 14 & 16.09 \\
      & Natural Science                & 14 & 16.09 \\
      & Office \& White Collar         & 14 & 16.09 \\
      & Finance \& Economics           & 9  & 10.34 \\
      & Mathematics \& OR              & 8  & 9.20 \\
      & Cybersecurity                  & 7  & 8.05 \\
      & Media \& Content Production    & 5  & 5.75 \\
    \midrule
    \multirow{6}{*}{ALFWorld}
      & Pick \& Place       & 35 & 25.00 \\
      & Examine in Light    & 13 & 9.29 \\
      & Clean \& Place      & 27 & 19.29 \\
      & Heat \& Place       & 16 & 11.43 \\
      & Cool \& Place       & 25 & 17.86 \\
      & Pick Two \& Place   & 24 & 17.14 \\
    \bottomrule
  \end{tabular}
\end{table}

\subsection{Hardware and Software Environment}

Experiments were orchestrated on a Linux server with two AMD EPYC 9454
48-core processors (96 physical cores and 192 logical CPUs) and 125~GiB of
system memory. The host ran Ubuntu 22.04.5 LTS with Linux kernel
6.8.0-124-generic. SkillsBench tasks were isolated in Docker containers and
were allocated the CPU, memory, storage, network, and timeout limits declared
by each task's public \texttt{task.toml}. ALFWorld was executed through its
TextWorld interface. The backbone LLM and text-embedding model were accessed
through OpenAI-compatible hosted APIs. The server was additionally equipped
with seven NVIDIA GeForce RTX 4090 D GPUs, each providing 49,140~MiB of
memory.

The Python environment was managed from a locked dependency file and used
Python 3.12.13. The principal package versions were ALFWorld 0.4.2, TextWorld
1.7.0, NumPy 2.4.3, SciPy 1.17.1, NetworkX 3.6.1, Pydantic 2.12.5, OpenAI
Python SDK 1.109.1, LiteLLM 1.80.0, and FastGraphRAG 0.0.5.

\subsection{Hyperparameters and Experimental Configuration}

All methods in a benchmark use the same task instances, skill library,
backbone model, and task verifier. The main comparison uses GPT-5.6 as the
agent backbone. SkillsBench uses a 200-skill library, while ALFWorld uses the
37 ALFWorld-specific skills contained in the same skill collection. Skill and
query embeddings are produced by \texttt{text-embedding-3-large} with 3,072
dimensions. Embeddings and the skill graph are constructed once and cached.
The cache is shared across queries but not across different skill libraries.

For graph construction, each skill considers eight linking candidates and a
dependency edge is retained at a matching threshold of 0.6. For GoS on
SkillsBench, four seed skills are selected and at most five skills are
returned. On ALFWorld, GoS uses five seeds and returns at most four skills in
the initial bundle. Personalized PageRank uses restart probability
$\alpha=0.2$, at most 50 iterations, and an $\ell_1$ convergence tolerance of
$10^{-6}$. In the GoS baseline implementation, reverse transitions are
weighted by relation type: 1.0 for dependency, 0.5 for workflow, 0.2 for
semantic, and 0.1 for alternative relations. SkillTrace instead uses the
reverse-aware dependency transition defined in the Method section.

The SkillTrace retrieval procedure follows the Method section. Here we report
only implementation and evaluation settings omitted from the main text. In
particular, SkillTrace uses the same PPR convergence settings reported above,
and its ALFWorld retrieval bundle is capped at six skills.

On ALFWorld, every method receives the same textual observation and action
grammar, domain randomization is disabled, the environment seed is 42, and an
episode is capped at 30 agent steps. On SkillsBench, one attempt is scheduled
for each task in the main comparison and the task-specific verifier determines
success. Infrastructure retries are limited to one and do not change the task
success rate. Invalid or incomplete trials are recorded as errors rather than
silently counted as successful completions.


\end{document}